\documentclass[runningheads]{llncs}
\usepackage{graphicx}
\usepackage{amsfonts}
\begin{document}
\title{A Machine Consciousness architecture based on Deep Learning and Gaussian Processes}
%
%
\author{Eduardo C. Garrido Merch\'an\inst{1}\and
Martin Molina\inst{2}}
%
%
\institute{Universidad Aut\'onoma de Madrid, Madrid, Spain
\email{eduardo.garrido@uam.es} \and
Universidad Polit\'ecnica de Madrid, Madrid, Spain
\email{martin.molina@upm.es}}
\titlerunning{Machine Consciousness based on Deep Learning and Gaussian Processes}
\maketitle              
\begin{abstract}
Recent developments in machine learning have pushed the tasks that machines can do outside the boundaries
of what was thought to be possible years ago. Methodologies such as deep learning or generative models have
achieved complex tasks such as generating art pictures or literature automatically. Machine Consciousness is a field that has been deeply studied and several
theories based in the functionalism philosophical theory like the global workspace theory have been proposed. In this work, we propose an architecture that may arise consciousness in a machine based in the global workspace theory and in the assumption that consciousness appear in machines that have cognitive processes and exhibit conscious behaviour. This architecture is based in processes that use the recent Deep Learning and generative process models. For every module of this architecture, we provide detailed explanations of the models involved and how they communicate with each other to create the cognitive architecture. We illustrate how we can optimize the architecture to generate social interactions between robots and genuine pieces of art, both features correlated with machine consciousness. As far as we know, this is the first machine consciousness architecture that use generative models and deep learning to exhibit conscious social behaviour and to retrieve pictures and other subjective content made by robots.

\end{abstract}
\section{Introduction}
Several reviews have been written about machine consciousness \cite{gamez2008progress} \cite{reggia2013rise} \cite{gamez2018human} that
try to sum up all the ideas that literature has proposed about the potential arisal of consciousness in machines \cite{chella2011artificial}. These ideas come from different areas such as artificial intelligence \cite{chella2007artificial}, neuroscience \cite{pennartz2019indicators} or philosophy \cite{searle2004mind}. Although consciousness can not be measured directly, there exist approaches that have provided potential measures of consciousness in machines \cite{arrabales2010consscale} \cite{raoult2015reviewing}.

Although the field generates controversy \cite{crick1994astonishing} as it lies in the margin of the scientific method, it has recently attracted the attention of relevant researchers of computer science such as Yoshua Bengio, who has provided an approach for how machine consciousness may arise with deep learning \cite{bengio2017consciousness}. As deep learning \cite{lecun2015deep} has generated machines that implement attention mechanisms \cite{graziano2017attention}, a new focus have emerged with the field of machine consciousness based in the astonishing hypothesis \cite{crick1994astonishing} that our intelligence and consciousness may arise from very simple principles.

Computational approaches for machine consciousness are based in the functionalism theory of consciousness \cite{reggia2013rise}. This theory claims that while mental states correspond to brain states, they are mental states due to their functionality, not due to their physical composition. Hence, consciousness may appear in machines that implement behaviors observed in humans that are correlated with consciousness.

Throughout the recent years, there has been amazing advances in the artificial intelligence and machine learning community \cite{murphy2012machine} that does not only include deep learning models. In the machine consciousness literature, it has been hypothesized that consciousness, or phenomenal states \cite{loar1990phenomenal}, may arise from machines that are able to perform tasks that humans are able to do when they are conscious \cite{damasio1999feeling} \cite{gamez2018human}. This is based in the hypothesis that if humans are conscious when producing complex behaviours, then,  machines may be conscious when they produce them too \cite{gamez2008development}.

We know, and have measured, that humans are conscious when performing these behaviours thanks to functional magnetic resonance imaging (fMRI) and related techniques \cite{he2009fmri} \cite{kamitani2005decoding}. These behaviours can include imagination \cite{stuart2007machine}, emotions \cite{shanahan2005consciousness}, language communication and social relations \cite{searle2002consciousness} or awareness of the environment \cite{kitamura2000can}.

Machine learning recent models are able to generate art \cite{elgammal2017can} that deviate from what they are fed to learn, are able to learn how to learn \cite{thrun2012learning}, learn from a few examples \cite{snell2017prototypical} and are able to transfer knowledge from a different task to behave better in a new one \cite{kaiser2017one}. The applications of these abilities include natural language generation \cite{deng2018deep}, understanding emotions \cite{cambria2016affective} or generating videos \cite{xiong2018learning}. We believe that if the philosophical theory that consciousness arises as a flux of information in any machine \cite{rose2006consciousness} is true, if we create a cognitive architecture \cite{chella2008cognitive} that is able to produce as many behaviours as possible that are correlated with consciousness in humans, then, the machine may as well arise, up to some extent, consciousness or phenomenal states.

We attempt to provide a bridge between the machine learning and the machine consciousness communities by providing the design of a cognitive architecture with machine consciousness behaviours through machine learning models. Several architectures have been proposed before \cite{dehaene2005ongoing} but none of them include both deep learning, generative processes and gaussian processes to generate interior cognitive processes and exterior behaviour and content. Section 2 will discuss related work. Then, in Section 3, we provide a detailed explanation of the modules of our architecture. Section 4 then provides the architecture that unifies these modules. We conclude our work with a section of conclusions and further work.

\section{Related Work}

Due to different theories explaining the origin of consciousness, several approaches have been proposed to tackle this problem. We first discuss the different processes involving machine consciousness \cite{gamez2008progress} and then, the different approaches that have tackled machine consciousness \cite{reggia2013rise}.

Machine consciousness processes involve mainly four categories ordered from 1 to 4 in function of how close to generating real awareness they are \cite{gamez2018human}. 

Level 1 includes machines that implements external behaviour associated with consciousness. Some of the described behaviours in the introduction section like social interactions implemented in machines would be level 1 and the field of artificial general intelligence \cite{goertzel2007artificial} lies in this level. Several authors \cite{harnad2003can} \cite{moor1988testing} argue that machines implementing these behaviours may produce consciousness, but there is controversy. Machines that implement cognitive characteristics like imagination \cite{aleksander2003axioms}, attention, emotion, depiction and planning are level 2 machines. When an architecture involving all these process exists, we are talking about level 3 machines, that is, machines with an architecture that is claimed to be a cause or correlate of human consciousness. Lastly, phenomenally conscious machines are level 4 machines based in the hypothesis that several level 2-3 design could emerge phenomenal states \cite{aleksander2007axiomatic}.

Several approaches have tackled the previous categories of machine consciousness. A classification of them all \cite{reggia2013rise} includes five categories: First one are methods based in the global workspace theory \cite{baars1997theatre}. According to this theory, consciousness emerges from a system, like the brain, with a collection of distributed specialized networks with a fleeting memory capacity whose focal contents are widely distributed to many unconscious specialized networks, called contexts. These contexts work together to jointly constrain conscious events and to shape conscious contents \cite{baars2007global}. These theory has support of the neuroscience community \cite{baars1994neurobiological} and the computer science community \cite{bengio2017consciousness}. We are also inspired by this theory to provide a cognitive architecture \cite{chella2008cognitive} with machine learning techniques. Other categories include methods that suggest that consciousness emerges from a certain amount of information processing and integration \cite{balduzzi2008integrated}, from creating an internal self-model \cite{perlis1997consciousness}, from generating higher-level representations \cite{aleksander1996impossible} and from attention mechanisms \cite{koch2007attention}.

Machine consciousness has risen as a research topic for the deep learning literature \cite{bengio2017consciousness}, where the interest resides in learning representations of high-level concepts of the kind humans manipulate with language. We suggest that machine learning and related techniques \cite{bahdanau2014neural} are able to work as a global workspace, process a high amount of information, can generate internal self-models and higher level representations and have attention mechanisms. Hence, machine learning and generative processes should be explored in this field.

\section{Machine Consciousness Correlated Processes}
We now provide the module design that implement cognitive processes and exhibit external behaviour that is correlated with consciousness \cite{gamez2018human}. In the selection of the cognitive processes to be simulated, we consider behaviors that can make an autonomous agent evolve to adapt to an unknown environment through observation and social interaction. These behaviours are also affected by processes that establish emotional connections between observed and imagined content (e.g., images generated by simulated dreams, emotion simulation, depiction of the environment) and that can be supported by novel techniques such as deep learning and generative methods.

\subsection{Simulating dreams} In order to simulate dreams, we first have to record photos $\mathbf{P}_i$ when being awake and store them in a semantic network \cite{sowa1987semantic}. Then, dreams will use that information $\mathbf{P} : \mathbf{P}_i \in \mathbf{P}$ to generate a sequence of images $\mathbf{D}_i \in \mathbf{D}$. We define a dream as a function $d$ that converts a subset of a sequence of images $\mathcal{P} \in \mathbf{P}$ and a subset of a sequence of style images $\mathcal{S} \in \mathbf{S}$ in a new set of images $\mathcal{D}$, that is $\mathcal{D} = d(\mathcal{P},\mathcal{S})$. In order to generate this procedure, we propose two processes for this simulator:

First, we classify images $\mathcal{P}$ into a semantic network $R$. We assume that a previous categorized semantic network $R$ exists and that a robot has already learned to classify images $\mathcal{P}$ into that network $R$.  An implementation of this process can have ImageNet \cite{deng2009imagenet} as semantic network. ImageNet is a resource with more than $14.000.000$ images $\mathcal{D}(\mathbf{X})$ and more than $21.000$ categories $\mathbf{y}$. ImageNet uses the hierarchy of WordNet \cite{fellbaum2012wordnet} to classify photos, having each category $y_i$ a semantic meaning and being organized as a graph $G = {V,E}$ that can be traversed, where $v\in V$ is the node representing category $y_i$. Convolutional neural networks \cite{krizhevsky2012imagenet} or advanced neural models as Efficient Net L2 \cite{xie2019self} or ResNet \cite{wu2019wider} neural models can classify photos into ImageNet. Let $NN$ be the neural model that implements the robot, the robot will classify each input image $P$ to category $y_i$, inserting it in the graph $G$ through the $NN$ trained on the ImageNet dataset $\mathcal{D}(\mathbf{X},\mathbf{y})$, that is: $y_i = NN(P|\mathcal{D}(\mathbf{X},\mathbf{y}))$.  

To feed images in the neural model $NN$ to be classified in the graph $G$, we need a robot with an integrated camara to take the photos $\mathbf{P}$ and define a period of being awake $T_a$ and asleep $T_s$. These parameters can be configured differently for every robot. We suggest to save additional images $\mathcal{S}$ that will represent different styles seen like for example dark places or broad landscapes in a different semantic network $R_s$.

\begin{figure}[htbp]
\centering{
\begin{tabular}{cc}
        \includegraphics[width=5.17cm]{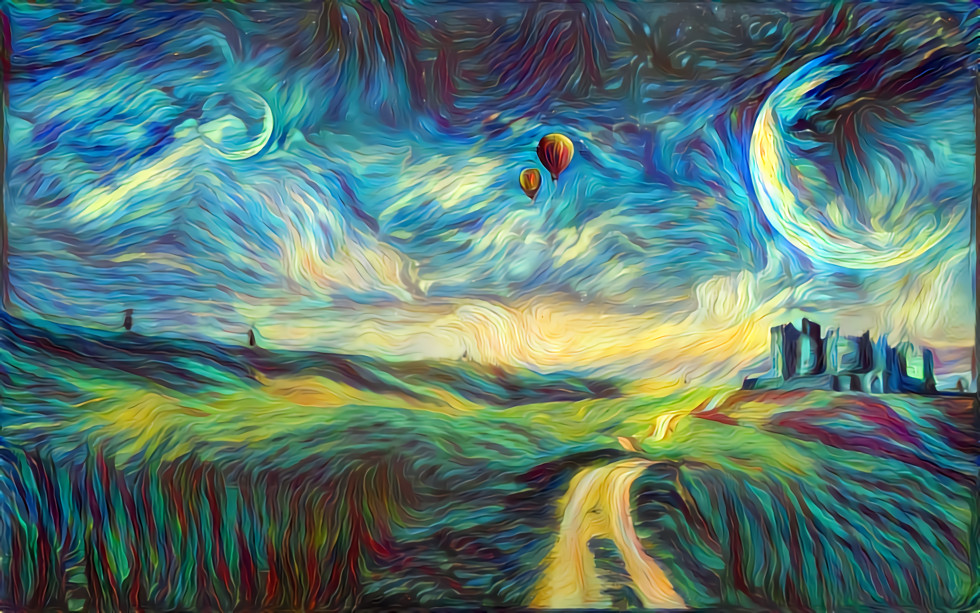} &
        \includegraphics[width=6cm]{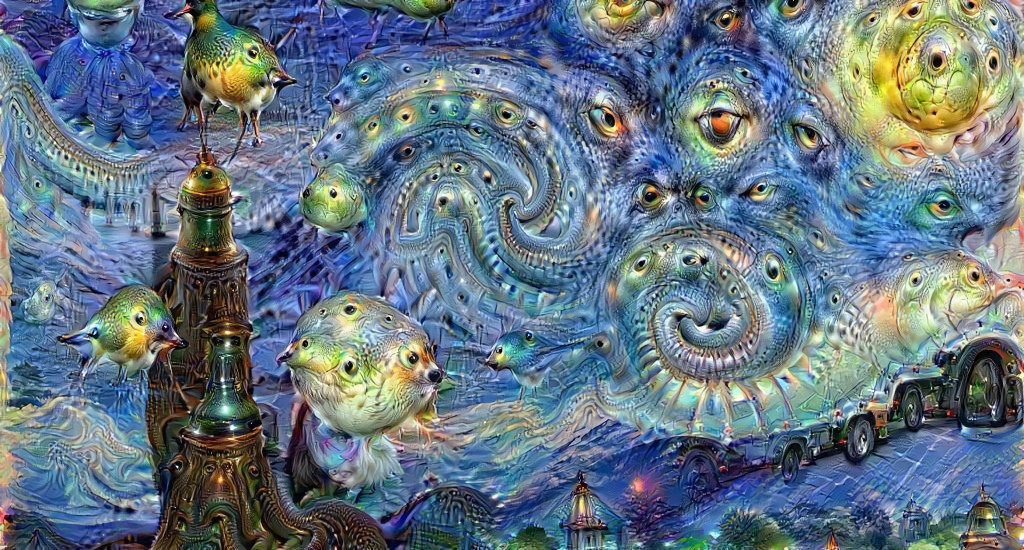}
\end{tabular}
}
\caption{Generated photos representing dreams by the Deep Dream Generator models (\url{http://deepdreamgenerator.com/)}}
\label{fig:dreams}
\end{figure}

Second, we need to define the dreaming state given by time $T_s$. We suggest to use a random walk \cite{roberts1997weak} like the one performed in the Metropolis Hastings algorithm \cite{chib1995understanding} to simulate movement into the semantic networks of images $R$ and styles $R_s$ that are related by semantic distance $d_s(y_i, y_j)$ in their graphs $G,G_s$ given by the number of edges that connect each category. At each step, we select two images $\mathbf{P}_i, \mathbf{P}_i^s$ and invoke Deep Style neural networks \cite{mordvintsev2015inceptionism} to generate a new image with the selected photo and an style applied $\mathbf{D}_i = DS_{nn}(\mathbf{P}_i, \mathbf{P}_s^i)$. Models such as a Generative Adversarial Network \cite{radford2015unsupervised} can be used. The robot will then attend the photo and save it. We can observe examples of generated photos using the Deep Dream Generator by this procedure in Figure \ref{fig:dreams}. 

Initial categories $y_{init}, y_{init}^s$ are chosen randomly. To select the new categories, we perform a random walk in the graph $G$ and $G_s$ given by some uniform distribution with a lower $l_l$ and upper $u_l$ limit, whose sampled value we will call step size $\omega$. If we set those parameters to a high number, dreams will contain different concepts and viceversa. We repeat the mentioned process by performing an iteration of the random walk. We store each generated image $\mathbf{D}_i$. The sequence of recreated images $\mathbf{D}$ recreates the dream. After dreaming, the robot will be awake, iterating both processes.

\subsection{Depiction. Being aware of the environment} We suggest to implement a robot that moves autonomously in a given environment $E$. For the sake of simplicity we are going to assume that $E$ is $2$-dimensional $E \in \mathbb{R}^2$. This robot will remember the images $\mathbf{D}$ that has previously dreamed as described in the previous module. The robot, when awake, will try to return to the location or neighbourhood $\mathcal{N} \in E$ where the images that has dreamed are located.

\begin{figure}[htbp]
\centering{
        \includegraphics[width=8cm]{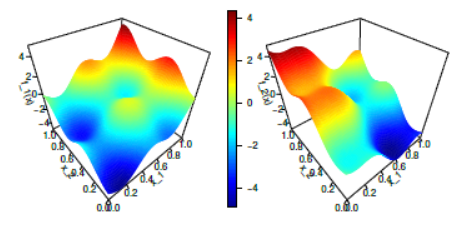}
}
\caption{Sampled functions from a 2-dimensional GP representing the importance value of each location of the environment for the robots.}
\label{fig:gps}
\end{figure}

We will generate a 2-dimensional function of location importance with a sample from a Gaussian Process  \cite{williams2006gaussian} $f_l^r \sim \mathcal{G}\mathcal{P}(0,k(\mathbf{x},\mathbf{x}')) \in \mathbb{R}^2$ over the environment $E$ , discretized by a grid, for each robot $r$ with interesting places to visit. We can observe examples of such functions at Figure \ref{fig:gps} The resolution of the grid $r_g$ can set the size of the environment $E$. Gaussian Processes models are flexible priors or
distributions over functions where inference takes place directly in the functional space $\mathcal{F}$. This functional space contains every possible environment that can be created $E \in \mathcal{F}$. 
\vspace{-.05cm}
\vspace{-.5cm}
The generated environment by the GP $f_l \in \mathbb{R}^2$ will contain high-valued locations of interest to take photos from and viceversa. When the robot reaches these places, it will take photos of the environment and save them for the dream module. Once visited, these places will be penalized by a local penalization procedure \cite{gonzalez2016batch}. These kind of procedures get a neighbourhood $N \in f_l$ centered in the place of interest $\mathbf{r} \in N$ and penalizes this zone by, for example, a multivariate gaussian distribution $f_l(N) = f_l(N) - MVG(\mathbf{r},\mathbf{I})$.

The robot will end navigation when it is exhausted after its time awake $T_a$. We can simulate fatigue through a non deterministic function $p(r,t) \in [0,1]$ of time since it has last slept. Each time that the robot takes a photo, fatigue will be incremented or decremented depending on the reward given by the photo. We can assign a threshold $\phi \in [0,1]$ for fatigue. When the robot is awake the threshold takes value $0$ and it is incremented as a function of time or by $\epsilon$ when taking a photo. The robot will fall asleep after the maximum time awake or if the non deterministic function samples a value higher than the threshold $p(r,t) > \phi$. The action of taking or not the photo $t_p$ in each place $g$ of the grid will be non deterministic and dependant on the value of $f_l$, that is $t_p(f_l,g)$. The robot can take a photo my sampling $t_p$ periodically after an amount of steps $s$ in $f_l$ Specific parametric functions can be configured for each robot.

The Gaussian process sample $f_l$ will be contaminated by i.i.d. gaussian noise $\epsilon \sim \mathcal{N}(0,\sigma_{gj})$ in each position of the grid $g$ and dimension $j$ in the period of being awake to favour exploration. Higher values of $\sigma$ will enforce exploration. The robot will navigate through the environment with a metaheuristic \cite{voss2012meta} (exploration-exploitation) or by the gradients of the Gaussian process \cite{solak2003derivative} (exploitation).  Random rewards will be put in the scenario. 

\subsection{Emotion simulation} In this section, we define a process that models emotions through objective functions $e(t) \in [0,1]$ of time. The main reason why we implement emotions in these robots is because they are going to influence the Gaussian Process prior $f_l$ of the environment $E$. If the robots feels confident and happy $e_h(t) \approx 1$, unknown near areas of $E$ to the position of the robot $g$ will be rewarded to be explored. If $R$ is a neighbourhood of $f_l$ containing a reward, we can reward its value by sampling from a multivariate gaussian distribution centered in the reward $f_l(R) = f_l(R) - MVG(\mathbf{r},\mathbf{I})$. By doing this process, the robot will enter a positive cycle and take photos of interesting places. By performing this action, we increment $e_h(t)$ by a uniform distribution which limits $[l,u]$ can be parametrized. If, in contrast, the robot feels sad and fear $e_h(t) \approx 0$, movement across the grid will be penalized by incrementing fatigue and decrementing the step size $\omega$ of the random walk, entering a negative loop. 

These cognitive processes will exhibit external behaviour that will show if a robot is happy or sad by its activity on the grid. We provide an exit of the cycles by images of dreams $\mathcal{D}$. Dreams can also influence emotions $e(t)$ and make the robot behave differently. If an image resembles a visited area that had got high value of $f_l$, happiness will be incremented by a parametrizable amount $e(t) = e(t-1) + \delta \sim U[l,u]$, where $t$ represents time. If images of places with low value of $f_l$ are displayed, the opposite operation will be performed $e(t) = e(t-1) - \delta \sim U[l,u]$. Happiness could also affect the fatigue function, by alleviating it if the robot is happy or increasing it in the other case.

Other emotions that may be optimized are curiosity and boredom $e_c(t) \in [0,1]$, that would affect the Gaussian Process sampled function $f_l$ by penalizing already saw places by an $MVG(\mathbf{0},\mathbf{I})$ and rewarding unknown places also by an $MVG(\mathbf{0},\mathbf{I})$. A last example can be friendship and solitude $e_f(t)$, based in relations with other robots that are going to be described further or courage and fear $e_c(t)$ that will condition the movements across the environment by incrementing the step size $\omega$ of the random walk. The described fatigue function can also be seen as an emotion. Particular parametric forms of the functions are open for the robot developer to be implemented.  

\subsection{Social relationships with other robots} If we want to simulate emotions $e(t)$ like the ones felt with humans to show behaviour correlated with consciousness, we need to model these emotions to be not only a function of the environment interaction $f_l$ but also of relationships with other robots. For this reason, we consider that an essential component for the cognitive processes of the robots must be the interaction with other robots to share experiences, in the form of photos $\mathcal{P}$ in this setting, and influence the emotions $e(t)$.

Emotions like friendship or solitude $e_s(t)$ are dependant on social interactions. We define here a social interaction $\alpha(\beta_x, \beta_y)$ as the change of a photo $\mathcal{P}_x$ of a robot $\beta_x$ with a photo $\mathcal{P}_y$ of a robot $\beta_y$ when both robots share the same location $g$ in the environment $E$.

Each robot $\beta_i$ has a different function sampled from the GP prior $f_l^i \sim \mathcal{G}\mathcal{P}(0,k(\mathbf{x},\mathbf{x}')) \in \mathbb{R}^2$ of the environment $E$. As each photo $\mathcal{P}_i$ related to a position of the grid $g_i$, it will have, for every robot $\beta_i$ a different value $f_l^i(g_i)$, conditioning the rest of the emotions. If the photo refers to a location that the robot likes according to its prior $f_l^i$, emotions will make the robot more active. Although, if this is not the case, the robot may enter a negative cycle.

By interacting with each other, robots $\mathbf{\beta}$ will share images $\mathcal{P}$ or dreamed images $\mathcal{D}$ of the environment $E$ that will modify their Gaussian Process sampled function $f_l^i$ and the other emotions of the robot. Specific parametric forms are again free for the programmer of the robot to be set. 

\section{An Unified Architecture for the Models}

In the previous section, we have described how can we implement behaviours correlated with consciousness in machines. All the described processes can be implemented in a certain amount of robots $\mathbf{\beta}$ with an environment $E$ that they can traverse and get photos $\mathcal{P}$ from. In this section, we provide a diagram with all the modules described to illustrate how the information flows in our architecture. 

Besides the processes described in the previous section and in order to be more general, the proposed architecture uses multimodal information (e.g., ambient music and texts in form of recipes, besides images). These processes would generate subjective creations, which can be correlated with their communication to processes generated in conscious states, such as recipe suggestions \cite{garrido2018suggesting} where in each position modelled by the GP the robot would find, with a probability sampled from a random variable, a suggestion of a recipe and generate in base of the recipe a degree of tastiness. Another alternative is to include ambient music simulations \cite{martinez2019simulating}, where in each position in the input space we would have an ambient noise sample, also with a probability distribution given by the sampling of a random variable, and the robot would have a ambient music simulator, that uses these samples to generate music, simulating imagination and conditioning the emotion simulator. 

\begin{figure}[htbp]
\centering{
        \includegraphics[width=12cm]{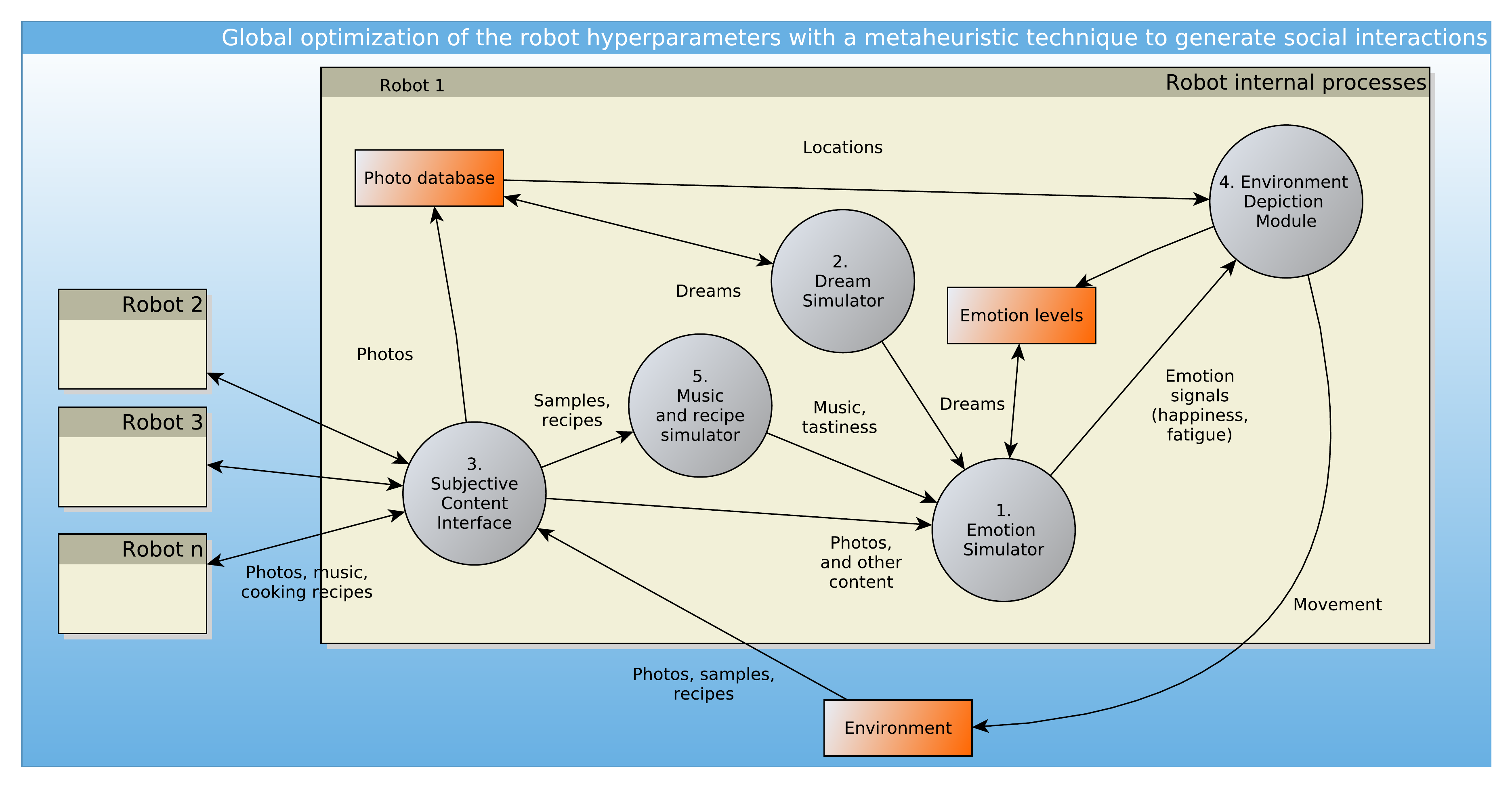}
}
\caption{Architecture of the proposed robots with behaviours correlated with consciousness. External processes that interact with the environment involve the depiction engine and the interface that collects data. Internal processes involve the dream, music and recipe generators and the emotion simulator, that condition the external behaviour. The hyperparameters of the robots models can be jointly optimized by a metaheuristic.}
\label{fig:diagram}
\end{figure}

All these processes generate the architecture that we can see in Figure \ref{fig:diagram}.
\vspace{-.5cm}
\vspace{-.5cm}
We can observe how robots share images and other information showing social behaviour. These interactions affect their emotions and incur in a different movement across the environment, reflecting emotions, commonly correlated with consciousness. Cognitive interior processes include dreaming images that are function of the perceived images and simulating music and cooking recipes, affecting emotions. These behaviours could, according to the cited theories, be a correlation of consciousness in robots.

Lastly, we propose the optimization of the different parametric forms $e(t)$ of the emotion engine and the parameters of the deep neural network for the dream simulation and the gaussian process for the environment in several simulations of the robots with a metaheuristic such as a genetic algorithm with the fitness being a function of the maximum number of interactions possible of the robots constrained to the maximum movement of the robots. By performing this optimization, we would end up having the optimum configuration for the robots to exhibit social behaviour, typically correlated with consciousness.

If the hyperparameters of the models are correctly optimized in future experiments, the outputs of these robots are hypothesized to be beautiful pieces of art in the form of pictures, music songs and cooking recipes, as all the environment can be seen as a reinforcement learning optimization technique to create subjective content and social interactions with conscious behaviours. Other architectures for machine consciousness just focus in cognitive processes and implementation of cited machine consciousness theories.

\section{Conclusions and further work}

We have described an architecture of processes that, if implemented in robots, exhibit external behaviour in the form of genuine art content and social behaviour. According to machine consciousness theory  \cite{gamez2018human}, both characteristics could be correlated with machine consciousness in robots \cite{gamez2008progress}. A significant novelty of this approach is the use of generative models based on the latest techniques of machine learning and deep learning to simulate processes such as imagination or depiction, where gaussian processes are flexible models that create functional spaces that contains lots of different environments. 

The presented architecture is a theoretical proposal that should be validated with practical tests. For this reason, we plan to implement all the processes in robots to get empirical evidence about the behaviour associated with consciousness and execute machine consciousness tests with natural language processing modules to verify if the robots are able to pass them. Further work will also include optimizing the emotions by some mechanism such as constrained Multi-objective Bayesian Optimization \cite{garrido2019predictive} in order to create a global and dynamical policy for the behaviour of the robots and including a weighted causal graph \cite{garridomerchn2020uncertainty} as knowledge base to generate more complex social relationships where even fake information could be shared or detected \cite{garridomerchn2020fake}. 

\section*{Acknowledgments}

The authors acknowledge the use of the facilities of Centro de Computaci\'on Cient\'ifica (CCC) at UAM and acknowledge financial support from Spanish
Plan Nacional I+D+i, grants TIN2016-76406-P and TEC2016-81900-REDT.

\bibliographystyle{acm}
\bibliography{machine_conciousness}
\end{document}